\documentclass[lettersize,journal]{IEEEtran}
\usepackage{amsmath,amsfonts}
\usepackage{algorithm}
\usepackage{array}
\usepackage[caption=false,font=footnotesize,labelfont=rm,textfont=rm]{subfig}
\usepackage{textcomp}
\usepackage{makecell}
\usepackage{stfloats}
\usepackage{url}
\usepackage{verbatim}
\usepackage{graphicx}
\usepackage{cite}
\usepackage{colortbl} 
\usepackage{hyperref} 
\usepackage{bigstrut}
\usepackage{caption}
\usepackage{amsmath}
\usepackage{multirow}                 
\usepackage{multicol}                 
\usepackage{multirow} 
\usepackage{siunitx}
\usepackage{pifont}       
\usepackage{bbding}    
\usepackage{flushend}
\usepackage{xcolor}  
\usepackage{colortbl}  
\usepackage{algpseudocode}
\usepackage{amssymb}

\usepackage{booktabs,makecell, multirow, tabularx}
\hypersetup{
    colorlinks=true, 
    linkcolor=blue, 
    citecolor=green, 
    urlcolor=blue 
}
\begin{document}

\title{

SwiTrack: Tri-State Switch for Cross-Modal Object Tracking
}

\author{Boyue Xu, \IEEEmembership{Student Member, IEEE}, Ruichao Hou, \IEEEmembership{Member, IEEE},  Tongwei Ren, \IEEEmembership{Member, IEEE} , Dongming Zhou, Gangshan Wu, \IEEEmembership{Member, IEEE} and Jinde Cao, \IEEEmembership{Fellow, IEEE}

\thanks{This work was supported by the National Natural Science Foundation of China (62072232, 62576098), the Key R\&D Project of Jiangsu Province (BE2022138), the Fundamental Research Funds for the Central Universities (021714380026), and the Collaborative Innovation Center of Novel Software Technology and Industrialization.
\emph{(Corresponding authors: Tongwei Ren)} }
\thanks{Boyue Xu, Ruichao Hou,  Tongwei Ren, Gangshan Wu are with the State Key Laboratory for Novel Software Technology, Nanjing University, Nanjing 210008, China (xuby@smail.nju.edu.cn; e-mail: rchou@nju.edu.cn;  rentw@nju.edu.cn; gswu@nju.edu.cn).}
\thanks{Dongming Zhou is with the School of Information Science and Engineering, Yunnan University, Kunming 650091, China (e-mail: zhoudm@ynu.edu.cn).}
\thanks{Jinde Cao is with the School of Mathematics, Southeast University, Nanjing 211189, China, and also with the Purple Mountain Laboratories, Nanjing 211111, China (email: jdcao@seu.edu.cn).}

\thanks{Manuscript received **** **, 2025; revised **** **, 2025.}
}

\markboth{Submitted to IEEE Journals/Transactions}%
{Shell \MakeLowercase{\textit{\emph{et al.}}}: A Sample Article Using IEEEtran.cls for IEEE Journals}


\maketitle

\begin{abstract}
Cross-modal object tracking (CMOT) is an emerging task that maintains target consistency while the video stream switches between different modalities, with only one modality available in each frame, mostly focusing on RGB-Near Infrared (RGB-NIR) tracking.
Existing methods typically connect parallel RGB and NIR branches to a shared backbone, which limits the comprehensive extraction of distinctive modality-specific features and fails to address the issue of object drift, especially in the presence of unreliable inputs.
In this paper, we propose SwiTrack, a novel state-switching framework that redefines CMOT through the deployment of three specialized streams. Specifically, RGB frames are processed by the visual encoder, while NIR frames undergo refinement via a NIR gated adapter coupled with the visual encoder to progressively calibrate shared latent space features, thereby yielding more robust cross-modal representations. For invalid modalities, a consistency trajectory prediction module leverages spatio-temporal cues to estimate target movement, ensuring robust tracking and mitigating drift.
Additionally, we incorporate dynamic template reconstruction to iteratively update template features and employ a similarity alignment loss to reinforce feature consistency. Experimental results on the latest benchmarks demonstrate that our tracker achieves state-of-the-art performance, boosting precision rate and success rate gains by 7.2\% and 4.3\%, respectively, while maintaining real-time tracking at 65 frames per second. Code and models are available at \href{https://github.com/xuboyue1999/mmtrack.git}{https://github.com/xuboyue1999/SwiTrack.git}.

\end{abstract}

\begin{IEEEkeywords}
Cross-modal tracking, state switch, visual adapter, near infrared, trajectory prediction.
\end{IEEEkeywords}

\begin{figure}[t]
\centering
  \includegraphics[width=0.49\textwidth]{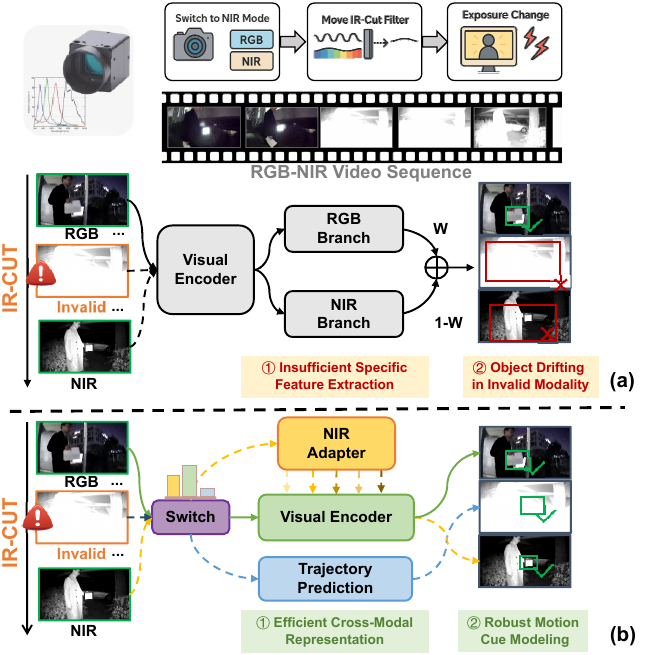}   
  \caption{Framework comparisons between existing and proposed cross-modal trackers. (a) Existing tracker: Utilizes symmetric dual branches following the visual encoder, which limits the extraction of modality-specific features and lacks robustness against modality invalidity due to over-exposure. (b) SwiTrack: Designs a tri-state switch for dynamic state assessment, a NIR gated adapter for feature modulation, and consistent trajectory prediction for robust motion cue modeling during switching.} 
  \label{fig:intro}  
\end{figure}


\IEEEPARstart{C}{ross-modal} object tracking (CMOT)~~\cite{cmotb} refers to tracking an object while the video stream alternates between different modalities, with only one modality active per frame. Leveraging a single sensor with a switchable optical filter, CMOT acquires multiple spectral bands without dual cameras, reducing hardware cost and avoiding image-alignment overhead. CMOT systems predominantly switch between RGB and near-infrared (NIR) bands~\cite {nearinfrared1,nearinfrared2}, because this pair can be realized simply by toggling a filter, enabling low-cost, all-weather, single-sensor tracking. The challenge lies in modeling disparate feature distributions within one network and staying robust under frequent switches. 

Recent CMOT approaches achieve promising results. As shown in Figure~\ref{fig:intro} (a), most approaches adopt a unified visual encoder to handle both RGB and NIR inputs. To capture modality-specific characteristics, they append simple branches after the shared encoder and use a modality classifier to generate weights or select the appropriate branch. This enables the tracker to apply tailored processing to different modalities while maintaining a unified backbone.
However, these trackers encounter two primary limitations: (1) neglecting physics-based imaging constraints. A shared backbone inadequately preserves modality-specific features, as later-stage branches fail to capture low and mid-level feature differences. (2) NIR imaging heavily relies on environmental infrared reflection. In scenarios where environmental reflection is low, an infrared illuminator is activated to enhance imaging. The abrupt activation, coupled with a wide camera aperture, causes over-exposure during the transition phase, leading to the unreliability or complete loss of visual features. Existing trackers fail to address this issue, leading to target drift due to invalid features, ultimately resulting in tracking failure.

To better handle modality differences and extract discriminative cross-modal features, we propose a novel tracker, SwiTrack, guided by a state-switching mechanism, as shown in Figure~\ref{fig:intro}(b). Although training separate models for each modality and switching between them can partially address the problem, our unified design eliminates the runtime overhead of model switching. Processing all modalities with a single model ensures that tracking information remains continuous rather than fragmented by repeated restarts, resulting in a more stable and reliable tracker. The state switch identifies the current input as RGB, NIR, or invalid. RGB inputs are directly processed by the visual encoder, while NIR inputs are further refined by a NIR gated adapter coupled with the visual encoder. The adapter adjusts NIR features by referencing dynamic template features. Since different feature levels require varying degrees of adaptation, we design a hierarchical gate that dynamically adjusts the weights for each layer. While deep semantic features require minimal modification, shallow features demand more substantial adjustments. Additionally, we introduce a similarity alignment loss during tracking to enhance dynamic template features, improving representational quality and tracking robustness.
To address invalid states, we introduce a consistent trajectory prediction (CTP) module that uses motion cues to maintain tracking when visual features are unreliable. Upon detecting an invalid state, the tracker freezes dynamic visual features and predicts the target position based on historical trajectories. Once valid input resumes, reliable tracking results are used to update the predictor. Unlike traditional motion models, our approach adaptively adjusts predictions by jointly considering modality and tracking reliability. Motion cues are also encoded in the unified encoder, helping to preserve robust tracking across dynamic cross-modal conditions.
Experimental results on the latest cross-modal tracking benchmarks
demonstrate that our method achieves state-of-the-art performance
improving PR and SR scores by 7.2\% and 4.3\%, respectively, while
achieving real-time tracking at 65 FPS.

In summary, we make three-fold contributions:
\begin{itemize}
\item We propose SwiTrack, a cross-modal tracker equipped with a state switch for adaptive tracking across RGB, NIR, and invalid states. 

  \item We propose the NIR gated adapter, which dynamically adjusts NIR features to ensure both NIR and RGB features are effectively extracted.
  \item We introduce a CTP module that exploits motion history to predict the target during invalid states and dynamically re-weights its filter using modality and tracking confidence, yielding robust trajectories under challenging conditions.
\end{itemize}
\section{Related Work}



\subsection{Cross-Modal Object Tracking}
CMOT does not require dual-modality aligned videos as input, thereby reducing data collection requirements and complexity. However, in CMOT, the modality shifts dynamically according to conditions, with only one modality active at a time. Such modality shifting poses significant challenges, as there exist substantial feature differences between near-infrared and RGB modalities, making it difficult for single-modality processing methods to bridge this gap. To address this issue, Li \emph{et al.}~\cite{marmot} introduced the first cross-modal tracking dataset, pioneering the definition of the task's challenges and characteristics while proposing a baseline approach that trains a classifier to recognize modality differences for modality distinction and applies different processing strategies accordingly. Liu \emph{et al.}~\cite{cmotb} further extended this work by increasing the dataset size, categorizing it into easy and hard subsets to better define the task's challenges, and proposing an end-to-end tracker with modality-aware representation learning, achieving promising results. ProtoTrack~\cite{prototype} introduced the concept of dynamic prototypes, differentiating templates across modalities, and adapting prototype selection based on varying conditions to enhance tracking performance. 

While these methods explore modality differences, they overlook challenges like over-exposure during modality shifting and underutilized motion cues that are modality-invariant. Additionally, most adopt a shared backbone with modality-specific branches, which limits the exploitation of modality-specific characteristics. In contrast, our SwiTrack explicitly models modality transitions and invalid states, enabling more robust and adaptive cross-modal tracking.

\subsection{Multi-Modal Object Tracking}
Multi-modal object tracking (MMOT) augments RGB with complementary signals, thermal, depth, or event streams, to sustain perception when the visible spectrum is unreliable~\cite{rgbtREVIEW,arkittrack,MMTRACK,rgbt1,PTM,amnet,hu2025exploiting}. Representative settings include RGB-T~\cite{apfnet,tbsi,gmmt}, RGB-D~\cite{DepthTrack,rgbd1k,arkittrack}, and RGB-E~\cite{swineft,HRCEU}.

A large number of works build MMOT by extending strong RGB trackers with two parallel branches and an explicit fusion stage. Under this paradigm, APFNet~\cite{apfnet} explores attribute-aware fusion to cope with diverse challenges; MTNet~\cite{mtnet} employs Transformers to model global cross-modal relations; and SPT~\cite{rgbd1k} applies Transformer-based extraction and fusion to better exploit complementary cues. While effective, parallel-branch designs usually increase computation and training complexity, and their reliance on simultaneous multi-stream inputs can complicate deployment and cross-modal transfer when one modality becomes invalid.

More recently, prompt-based approaches reduce overhead by injecting lightweight modules into powerful RGB foundations~\cite{VIPT,protrack,untrack}. ProTrack~\cite{protrack} is a pioneering unified framework that adapts a base tracker across RGB-T/RGB-D/RGB-E tasks via simple adapters. ViPT~\cite{VIPT} leverages prompt-driven adapters and a Fovea-Fusion mechanism to improve cross-modal integration, while SDSTrack~\cite{sdstrack} introduces robust training to mitigate low-quality modalities. Beyond purely visual cues, OneTracker~\cite{onetracker} incorporates textual prompts to enrich context. In parallel, UN-Track~\cite{untrack} and SUTrack~\cite{sutrack} pursue scalable unification over a broad spectrum of single- and multi-modal tracking tasks.

Although MMOT also deals with multiple modalities, it typically relies on paired multi-modal images and places more focus on modality fusion. While the use of adapters in these works offers valuable insights, they still fall short in addressing the challenges of dynamic modality shifting, which is critical in cross-modal tracking tasks.


\begin{figure*}[t]
  \centering
  \includegraphics[width=0.99\textwidth]{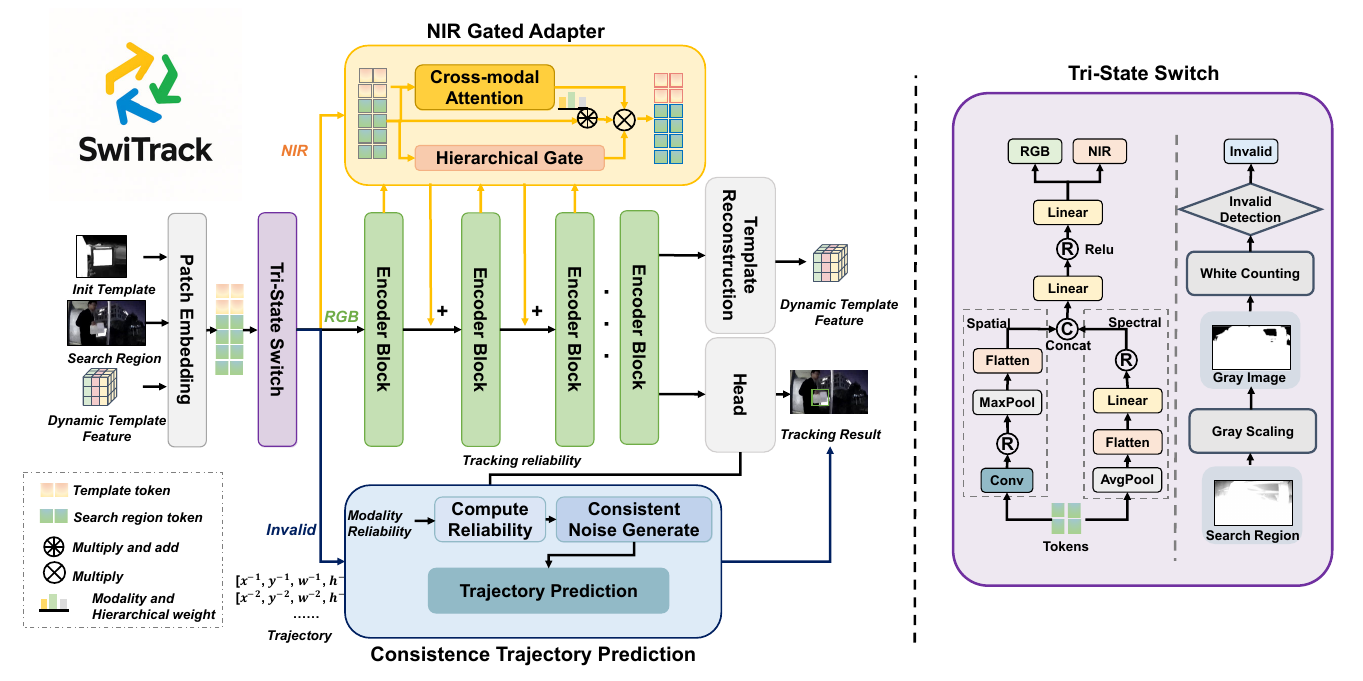}   
  \caption{The framework of the proposed SwiTrack. Templates, the search region, dynamic template features, and trajectory are tokenized and processed by an $l$-layer ViT block. It identifies the search region’s state, RGB, NIR, or invalid, and activates the NIR gated adapter for modality adaptation. The template reconstruction module updates dynamic templates using current visual features, while the prediction head generates tracking results. When the state is invalid, the consistent trajectory prediction module estimates the trajectory without relying on visual features. State switch includes modality change detection based on features and invalid detection based on the image.}   
  \label{fig:pipeline}  
\end{figure*}
\subsection{Visual Object Tracking}
Single-modal object tracking methods mostly rely on RGB. Existing approaches can be broadly categorized into three types: classification-based, Siamese-based, and Transformer-based methods.
Classification-based methods~\cite{dimp,atom} treat tracking as a foreground-background classification task. MDNet~\cite{mdnet} continuously updates its classifier with online samples to enhance discrimination. RT-MDNet~\cite{rtmdnet} improves efficiency, while LTMU~\cite{ltmu} and MIRNet~\cite{mirnet} enhance robustness by integrating temporal and attention mechanisms. However, these methods often suffer from slow tracking speeds, limiting their real-time applicability.

Siamese-based methods~\cite{siam1,siam2,siam3}, such as SiamFC~\cite{siamfc}, perform tracking by comparing feature similarity between a template and the search region. Follow-ups explore stronger heads and inference schemes (\emph{e.g.}, anchor-free scoring and improved feature fusion) while keeping the twin-branch paradigm. LightTrack~\cite{lighttrack} further leverages one-shot neural architecture search to obtain a compact backbone/head that maintains accuracy and achieves real-time speed on edge devices. Despite these advances, Siamese trackers still rely largely on static templates with limited online adaptation, making them vulnerable to large appearance changes, long sequences, and dramatic pose or illumination shifts.

Transformer-based methods~\cite{transstrack1} have recently gained popularity due to their superior capacity for modeling both spatial and temporal dependencies. OSTrack~\cite{ostrack} introduces a unified one-stream framework leveraging self-attention. SwinTrack~\cite{swintrack} improves hierarchical feature extraction using Swin Transformer~\cite{swintrans}. SeqTrack~\cite{seqtrack,seqtrackv2} formulates tracking as a sequential prediction task, while ARTrack~\cite{ARTRACK,artrackv2} further unifies appearance and temporal cues for enhanced robustness and accuracy. Some methods use token context–aware learning that automatically mines high-quality reference tokens for efficient Transformer tracking~\cite{lmtrack}, and auto-regressive sequential pretraining that models temporal dynamics via generative objectives before fine-tuning for tracking~\cite{arp}. In parallel, self-supervised trackers reduce label cost by learning from raw videos~\cite{SSTrack}.

Single-modal object tracking has achieved impressive results and requires only single-modality input. However, they are typically designed for a fixed modality and struggle when modalities switch dynamically, as they cannot extract modality-specific features. In contrast, our method adapts to different modality states by employing tailored strategies for each, enabling more robust and effective cross-modal tracking.




\section{Method}
\subsection{Overview}
The framework of the SwiTrack is illustrated in Figure~\ref{fig:pipeline}, and consists of several key components: patch embedding, ViT encoder block, state switch, NIR gated adapter, CTP, template reconstruction, and prediction head. The inputs include the search region, initial template, dynamic template features, and trajectory, all of which are converted into tokens via the patch embedding layer. The receiver receives both the tokenized search region and its raw image to determine whether the state is RGB, NIR, or invalid. If the state is NIR, the NIR gated adapter is activated to refine the NIR with the encoder block to progressively adjust shared latent space features for enhanced cross-modal representation. After encoding, the prediction head generates the tracking results and confidence scores, while the template reconstruction module updates the dynamic template features. In cases of invalid, where visual features become unreliable, the CTP module predicts the target’s position using historical trajectory data, preventing drift and ensuring stable tracking.

\subsection{Tir-State Switch}
Determining when a state switch occurs and when an invalid state causes feature degradation accurately is one of the core challenges in cross-modal tracking. Only by precisely identifying the timing of these switches can appropriate processing strategies be applied. Existing modality classification methods~\cite{cmotb,marmot} typically focus solely on distinguishing between RGB and NIR, neglecting the transitional phase where overexposure occurs due to illumination changes, which can lead to target drift when visual features disappear. Moreover, most existing approaches rely on simple classifiers for state identification, failing to effectively account for modality differences in cross-modal tracking.

As illustrated on the left of Figure.~\ref{fig:pipeline}, this module not only accurately classifies RGB and NIR but also identifies invalid states caused by over-exposure. The left part is responsible for RGB and NIR classification, taking the current search region features as input and processing them in two ways: one focuses on spatial differences between modalities, while the other distinguishes them at the spectral level. This process can be formulated as follows:
\begin{align}
\mathbf{F}_{spa} &= \mathrm{Flatten}\Big(\mathrm{MaxPool}\big(\mathrm{Relu}\left( \mathrm{Conv}(\mathbf{F}_{in} )\right)\big)\Big), \label{eq1} \\
\mathbf{F}_{spe} &= \mathrm{Relu}\Big(\mathrm{Linear}\big(\mathrm{Flatten}\left( \mathrm{AvgPool}(\mathbf{F}_{in} )\right)\big)\Big), \label{eq2}
\end{align}
where $\mathbf{F}_{in}$, $\mathbf{F}_{spa}$ and $\mathbf{F}_{spe}$ are the input search region feature, spatial branch feature and spectral branch feature; respectively, $\mathrm{Flatten(\cdot)}$ represents a flatten operation, $\mathrm{MaxPool(\cdot)}$ and $\mathrm{AvgPool(\cdot)}$ represent adaptive max-pooling and adaptive average-pooling, receptively.  
After extracting the two sets of features, we concatenate them and pass the combined representation through two linear layers to obtain the current modality probabilities:
\begin{align}
\mathbf{W}_{m} &=
\sigma\!\Bigl(
  \mathrm{Linear}\bigl(
    \mathrm{ReLU}\!\bigl(
      \mathrm{Linear}([\mathbf{F}_{\text{spa}},\,\mathbf{F}_{\text{spe}}])
    \bigr)
  \bigr)
\Bigr), \label{eq1}
\end{align}
where $\sigma(\cdot)$ represents the sigmoid function and $\mathbf{W}_{m}$ represents the modality weight.
Additionally, we incorporate overexposure detection. We observe that when over-exposure occurs, the image is predominantly covered by white regions, making pixel-based detection simpler and more accurate than feature-based methods. Therefore, we design a pixel-based over-exposure detector, as shown on the right side of Figure ~\ref{fig:pipeline}. The input to this module is the raw search region image, which is first converted to grayscale. We then count the number of white pixels, and if the number of white pixels exceeds a predefined threshold, the region is classified as overexposed.




\subsection{NIR Gated Adapter}
\begin{figure}[t]
\centering
  \includegraphics[width=0.45\textwidth]{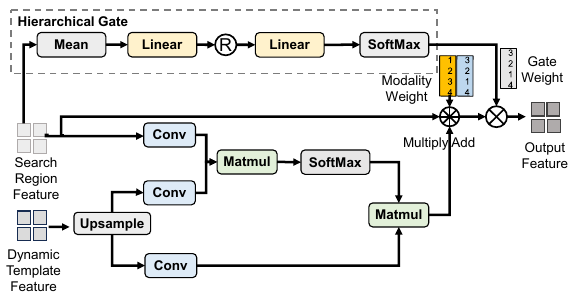}   
  \caption{Detailed design of the NIR gated adapter, which aims to align the NIR modality with the original features after state switching.} 
  \label{fig:adapter}  
\end{figure}
To adapt the features before and after state switching, we design a NIR gated adapter, as shown in Figure~\ref{fig:adapter}. Unlike other methods~\cite{cmotb} that employ a dual-branch design, our approach does not rely on two separate modality-specific branches. Instead, we introduce a gated adapter that is activated only when the state switches to NIR, otherwise, it simply outputs the original features. Additionally, different levels of features require varying degrees of adaptation: shallow features contain more texture details and thus require greater transformation, whereas deep features encode abstract semantic information and need minimal adjustment. To account for this, we incorporate a hierarchical gating mechanism that dynamically adjusts the weight of the adapter at each layer.
\begin{align}
\mathbf{W}_{i} &= \mathrm{SoftMax}\Big(\mathrm{Linear}\big(\mathrm{Relu}\left( \mathrm{Linear}\left(\mathrm{Mean}(\mathbf{F}_{sr}) \right)\right)\big)\Big), \label{eq1} 
\end{align}
where $\mathbf{F}_{sr}$ represents the search region feature and $\mathbf{W}_{i}$ represents the i-layer adapter gate weight.
The input to the adapter consists of the current search region features and the dynamic template features, whose purpose is to extract the core features from the previous dynamic template when the state switch occurs, ensuring that the current features remain more consistent with those before the switch. This is primarily achieved through attention:
\begin{align}
\mathbf{F}^{'}_{sr} &= \mathrm{Softmax}\left(\frac{\mathbf{Q}\cdot \mathbf{K}^{T}}{\sqrt{d_{k}}}\right) \cdot \mathbf{V},\\
\mathbf{F}_{o} &=\mathbf{W}_{i}\left( \mathbf{W}_{m}\otimes\mathbf{F}^{'}_{sr} \oplus (1-\mathbf{W}_{m})\otimes\mathbf{F}_{sr}\right),
\end{align}
where $\mathbf{F}_{sr}$, and $\mathbf{F}^{'}_{sr}$ represent the search region feature and enhanced search region feature, respectively, $\mathbf{W}_{i}$ is the same in equation(4).

In this way, the NIR gated adapter is only activated when the input state is NIR, adjusting NIR search region features based on the existing modality features. This adaptation makes the NIR features more compatible with the RGB, ultimately improving the tracking robustness.
\begin{algorithm}[t]
\caption{Consistent Trajectory Prediction (CTP)}
\label{alg:ctp}
\begin{algorithmic}[1]
\Require frames $\{I_t\}_{t=0}^{T}$, initial box $b_0$
\Ensure predicted boxes $\{\hat b_t\}_{t=0}^{T}$

\State $\mathbf{x}_0 \gets \text{box2state}(b_0)$
\State covariances $\mathbf{P}_0,\mathbf{Q}_0,\mathbf{R}_0$

\For{$t = 1,\dots,T$}
    \State $(x_t,\alpha_t) \gets \text{sampleTarget}(I_t,\hat b_{t-1})$
    \State $(\hat b^{\text{vis}}_t, s_t) \gets \text{appearanceTracker}(x_t)$
    \State $m_t \gets \text{modalityScore}(I_t)$
    \State $\text{inv}_t \gets \text{isOverExposed}(I_t)$

    \If{$\text{inv}_t$}
        \State \Comment{invalid: prediction-only with inflated process noise}
        \State $\mathbf{Q} \gets 1.5\,\mathbf{Q}$
        \State \textsc{CTP-Predict}$(\mathbf{x}_{t-1},\mathbf{P}_{t-1};\,\mathbf{Q}) \to (\mathbf{x}_t,\mathbf{P}_t)$
    \Else
        \State \Comment{valid: reliability-aware correction + prediction}
        \State $r_t \gets \max(\varepsilon,\; s_t\cdot|2m_t-1|)$
        \State $\mathbf{R} \gets \mathbf{R}_0 / r_t$
        \State $\mathbf{z}_t \gets \text{box2state}(\hat b^{\text{vis}}_t)$
        \State \textsc{CTP-Update}$(\mathbf{x}_{t-1},\mathbf{P}_{t-1};\,\mathbf{z}_t,\mathbf{R}) \to (\tilde{\mathbf{x}}_t,\tilde{\mathbf{P}}_t)$
        \State \textsc{CTP-Predict}$(\tilde{\mathbf{x}}_t,\tilde{\mathbf{P}}_t; \mathbf{Q}) \to (\mathbf{x}_t,\mathbf{P}_t)$
    \EndIf

    \State $\hat b_t \gets \text{state2box}(\mathbf{x}_t)$
    \State $\hat b_t \gets \text{clipBox}(\hat b_t,I_t)$
\EndFor
\end{algorithmic}
\end{algorithm}

\subsection{Consistence Trajectory Prediction}
\begin{figure}[t]
\centering
  \includegraphics[width=0.47\textwidth]{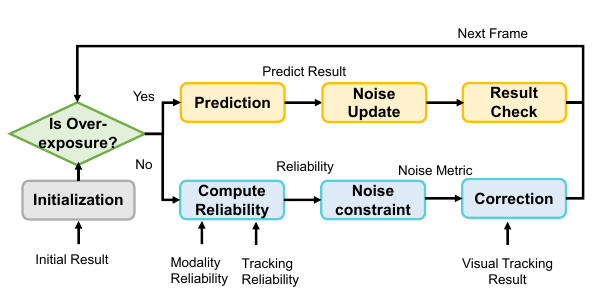}   
  \caption{Detailed design of the consistency trajectory prediction, which aims to predict the tracking trajectory when over-exposure occurs.} 
  \label{fig:motion}  
\end{figure}

To maintain tracking accuracy during invalid modality states, we introduce the CTP module. This module leverages motion cues to predict the target’s trajectory when visual features are unreliable.
As illustrated in Figure~\ref{fig:motion}, the proposed CTP module begins with an initialization phase at the start of each sequence. This includes setting the initial state vector $\mathbf{x}_{0}$ based on the target's initial position, the state covariance matrix $\mathbf{P}_0$ to represent uncertainty, the measurement noise covariance matrix $\mathbf{R}_0$, and the process noise covariance matrix $\mathbf{Q}_0$.
During valid input, the prediction model is corrected using the current visual tracking results. To ensure robustness, we first compute a reliability score for the current observation by jointly considering the modality reliability inferred from the state switch and the tracking confidence from the prediction head:
\begin{align}
r=\max\left(\epsilon,s\cdot|2m-1|\right), \label{eq1} 
\end{align}
where $s$ represents the current tracking reliability score, $\epsilon$ is 1e-3 and $m$ represents the modality reliability score. When $m$ approaches 0 or 1, the tracker confidently identifies the input as RGB or NIR, respectively. However, when $m$ is close to 0.5, it indicates uncertainty in the input state, typically occurring during state switching. In such cases, the reliability of the current observation should be down-weighted. This combined confidence is then used to guide the correction step of the current frame in the motion estimation.
\begin{align}
\mathbf{S} &= \mathbf{H}_{t}\mathbf{P}_{t|t-1}\mathbf{H}_{t}^{T}+\mathbf{R}_{t}/r, \label{eq2}\\
\mathbf{K}_{t} &= \mathbf{P}_{t|t-1}\mathbf{H}_{t}^{T}\mathbf{S}^{-1}, \label{eq3}\\
\mathbf{x}_{t|t} &= \mathbf{x}_{t|t-1}+\mathbf{K}_{t}\mathbf{y}_{t}, \label{eq4}\\
\mathbf{P}_{t|t} &= (\mathbf{I}-\mathbf{K}_{t}\mathbf{H}_{t})\mathbf{P}_{t|t-1}, \label{eq5}
\end{align}
where $\mathbf{x}_{t|t-1}$ and $\mathbf{x}_{t|t}$ denote the predicted state at time t before and after the update step, respectively; $\mathbf{y}_{t}$ represents the difference between the predicted and the measurement; $\mathbf{K}_{t}$ is the Kalman gain $\mathbf{S}$ is the innovation covariance matrix; $\mathbf{R}_{t}/r$ is the observation noise covariance matrix; adaptively adjusted based on the computed reliability score; $\mathbf{I}$ denotes the identity matrix.
\begin{align}
\mathbf{x}_{t|t-1} &= \mathbf{F}\mathbf{x}_{t-1|t-1}, \label{eq1} \\
\mathbf{P}_{t|t-1} &= \mathbf{F}\mathbf{P}_{t-1|t-1}\mathbf{F}^{T}+\mathbf{Q}, \label{eq2}\\
\mathbf{Q} &=\theta \cdot \mathbf{Q},\label{eq3}
\end{align}
where $\mathbf{F}$ is the state transition matrix, and $\mathbf{Q}$ represents the process noise covariance matrix, which increases over consecutive predictions to reflect the growing uncertainty during continuous estimation without reliable observations. In this way, guided by both modality reliability and tracking reliability, the proposed module can effectively predict the target position during state invalidity, thereby preventing target loss when visual features become unreliable. The pseudocode of the CTP module is provided in Algorithm~\ref{alg:ctp}.

We then validate the predicted target position to ensure it falls within a reasonable range and incorporate the prediction into the historical sequence to facilitate the next frame's estimation.
\begin{table*}
    \centering
    \setlength{\tabcolsep}{4pt}
    \small
    \caption{Comparison between the proposed SwiTrack and the SOTA trackers on different subsets. The best results are highlighted in \textbf{bold}. The performance is evaluated in terms of Precision Rate (PR) and Success Rate (SR).}
    \label{tablecompare}
    \begin{tabular}{c|c|c|cc|cc|cc|c}
        \toprule
        &\multirow{2}{*}{Methods} &\multirow{2}{*}{Years}& \multicolumn{2}{c|}{Easy Set} & \multicolumn{2}{c|}{Hard Set}& \multicolumn{2}{c|}{Joint Set} &\multirow{2}{*}{FPS$\uparrow$}\\
        && & PR$\uparrow$ & SR $\uparrow$ & PR $\uparrow$& SR$\uparrow$& PR$\uparrow$ & SR $\uparrow$&\\
        \midrule
        \multirow{10}{*}{\rotatebox{90}{Single}}   
        &TransT~\cite{transt}&$\text{CVPR21}$ & 58.0 & 51.0 & 29.6 & 41.2 & 44.1 & 47.0 & 50 \\
        &OSTrack~\cite{ostrack}&$\text{ECCV22}$ & 51.0 & 46.2 & 28.3 & 39.6 & 40.7 & 43.2 & 93 \\
        &AiATrack~\cite{aiatrack}&$\text{ECCV22}$ & 62.1 & 54.9 & 31.7 & 45.6 & 48.3 & 50.7 & 38 \\        
        &ARTrack~
        \cite{ARTRACK}&$\text{CVPR23}$ & 59.7 & 53.1 & 34.3 & 44.8 & 48.2 & 49.4 & 26 \\
        &SeqTrack~\cite{seqtrack}&$\text{CVPR23}$ & 57.9 & 51.0 & 31.8 & 43.3 & 46.1 & 47.5 & 40 \\ 
        &DropTrack~\cite{dropTRACK}&$\text{CVPR23}$ & 60.6 & 53.8 & 35.9 & 46.8 & 49.4 & 50.6 & 30 \\
        &MixFormerV2~\cite{mixformerv2}&$\text{NIPS23}$ & 52.2 & 47.8 & 28.5 & 39.8 & 41.5 & 44.2 & 165 \\
        &GRM~\cite{GRM}&$\text{CVPR23}$ & 50.5 & 46.0 & 29.8 & 40.4 & 41.1 & 43.4 & 45 \\
        &ROMTrack~\cite{ROMTRACK}&$\text{ICCV23}$ & 55.2 & 49.5 & 31.6 & 42.0 & 44.5 & 46.1 & 62 \\
        &CompressTrack~\cite{compresstrack}&$\text{ICCV25}$ & 44.7 & 41.5 & 23.5 & 35.6 & 35.1 & 38.9 & \textbf{250} \\
        &SSTrack~\cite{SSTrack}&$\text{AAAI25}$ & 60.2 & 51.0 & 31.0 & 42.2 & 47.0 & 47.0 & 60 \\
        \midrule
        
        \multirow{6}{*}{\rotatebox{90}{Multi}} 
        &VIPT~\cite{VIPT}&$\text{CVPR23}$ & 44.8 & 41.2 & 21.7 & 33.4 & 34.3 & 37.7 & 36 \\
        &SDSTrack~\cite{sdstrack}&$\text{CVPR24}$ & 43.4 & 39.4 & 22.4 & 34.3 & 33.9 & 37.1 & - \\
        &UNTrack~\cite{untrack}&$\text{CVPR24}$ & 53.7 & 48.1 & 28.1 & 39.4 & 48.1 & 44.1 & - \\
        &BAT~\cite{bat}&$\text{AAAI24}$ & 48.5 & 43.3 & 26.1 & 36.9 & 38.3 & 40.4 & - \\
        &SUTrack~\cite{sutrack}&$\text{AAAI25}$ & 59.1 & 53.2 & 32.1 & 44.2 & 46.9 & 49.1 & - \\
        &XTrack~\cite{xtrack}&$\text{ICCV25}$ & 55.5 & 49.5 & 29.3 & 40.8 & 43.6 & 45.6 & 15 \\
        \midrule
        
        \multirow{5}{*}{\rotatebox{90}{Cross}} 
        &MArMOT$_{r}$~\cite{marmot}&$\text{AAAI22}$ & 53.0 & 43.7 & 16.2 & 23.5 & 35.6 & 53.3 & 23 \\
        &MArMOT$_{d}$~\cite{marmot}&$\text{AAAI22}$ & 73.9 & 63.6 & 30.6 & 46.3 & 53.8 & 54.6 & 33 \\
        &MAFNet$_r$~\cite{cmotb}&$\text{TCSVT24}$ & 53.0 & 43.7 & 16.2 & 23.5 & 35.6 & 33.8 & 23 \\
        &MAFNet$_d$~\cite{cmotb}&$\text{TCSVT24}$ & 74.4 & 63.8 & 32.2 & 47.0 & 55.1 & 56.1 & 36 \\
         \rowcolor{gray!20} \cellcolor{white}&\textbf{Ours}&- & \textbf{77.6} & \textbf{65.9} & \textbf{41.3} & \textbf{51.5} & \textbf{62.3} & \textbf{60.4} & 65 \\ 
        \bottomrule
    \end{tabular}
\end{table*}
\subsection{Template Reconstruction}
To prevent the tracker from over-relying on the initial template and to mitigate feature discrepancies caused by state switching, we follow the baseline~\cite{artrackv2} and use the template reconstruction module. This module reconstructs dynamic template features based on the current tracking results, which are then used alongside the initial template during tracking.

The structure of the template reconstruction module remains consistent with the baseline, utilizing an 8-layer Vision Transformer. It takes the dynamic template from the previous frame as input and refines it accordingly. During training, the reconstructed template is compared with the ground truth target template of the current frame. Given the modality differences introduced by state switching, we assume that the reconstructed template does not need to be pixel-wise identical to the ground truth but should instead maintain feature similarity. Therefore, we replace the original mean squared error loss with a similarity error, ensuring better reconstruction performance.
\subsection{Modality Consistent Loss}
Our prediction head follows the baseline, employing a multi-layer MLP to predict the current target position and confidence score.

Regarding the modality consistent loss (MCL), our design consists of three components: tracking loss, modality loss, and template reconstruction similarity loss. The tracking loss aligns with most trackers, comprising an L1 loss for classification and an SIoU loss for regression. Additionally, we incorporate a gradually decaying CE loss to accelerate convergence in the early training stages:
\begin{equation}
\begin{aligned}
\mathcal{L}_{tra} = \lambda_1 \mathcal{L}_1\  + \lambda_2 \mathcal{L}_{SIoU} +\lambda_3\frac{N-C}{N}\mathcal{L}_{ce} ,
\end{aligned}
\end{equation}
where $C$ represents the current epoch number and $N$ represents total epoch, $\hat{b_t}$ is the corresponding ground truth,$\lambda_1$, $\lambda_2$ and $\lambda_3$ are set 5, 2 and 2, respectively. To compensate for the loss of modality, we utilize binary cross-entropy loss to constrain the state switch during training, thereby ensuring an accurate prediction of state changes.
\begin{equation}
\begin{aligned}
\mathcal{L}_{mod} &=  \alpha \mathcal{L}_{bce}\left ( m_t,\hat{m_t}  \right ) ,
\end{aligned}
\end{equation}
where $m_t$, $\hat{m_t}$ represents the ground-truth modality and predicted modality, respectively, $\alpha$ is set to 2. For the template reconstruction similarity loss, we adopt a cosine similarity loss to encourage the reconstructed template features to be as similar as possible to the ground truth features. This loss also decays gradually during training, ensuring that in later stages, the tracker relies more on existing features for target localization rather than continuously refining the reconstructed template:
\begin{equation}
\begin{aligned}
\mathcal{L}_{tem} &=  \zeta \frac{N-C}{N}\mathcal{L}_{sim}\left ( f_t,\hat{f_t}  \right ) ,
\end{aligned}
\end{equation}
where $\mathcal{L}_{sim}$ represents cosine similarity function, $f_t$, $\hat{f_t}$ represents reconstruction feature and template feature, respectively, $ \zeta$ is set 2.
The total loss is the sum of these three components:
\begin{equation}
\begin{aligned}
\mathcal{L} &=\mathcal{L}_{tra}+ \mathcal{L}_{mod} +\mathcal{L}_{tem}.
\end{aligned}
\end{equation}
\section{Experiment}
\subsection{Dataset and Evaluation}

To comprehensively evaluate our method, we evaluate SwiTrack on the Cross-Modal Object Tracking Benchmark CMOTB~\cite{cmotb}. CMOTB comprises 1,000 cross-modal video sequences, each containing at least one modality switch.  
The benchmark is split into an easy subset that covers common tracking scenarios, a hard subset designed to stress cross-modal challenges such as modality delay, and the combined joint set for overall assessment.

Following prior work, we adopt two metrics:

\textbf{Precision Rate (PR).}  
PR measures centre-location accuracy:
\begin{equation}
  \text{PR} = \frac{N_p}{N} \times 100\%,
\end{equation}
where $N_p$ is the number of frames whose centre-location error (CLE) is below 20 pixels, and $N$ is the total number of frames.

\textbf{Success Rate (SR).}  
SR evaluates bounding-box overlap:
\begin{equation}
  \text{SR} = \frac{N_s}{N} \times 100\%,
\end{equation}
where $N_s$ denotes the number of frames whose intersection-over-union (IoU) with ground truth exceeds 0.5.

These definitions follow the original CMOTB and are widely used in recent cross-modal tracking studies.


\subsection{Implementation Details}
The proposed method is trained on a server with a 5.2GHz CPU and a RTX 4090 GPU with 24GB of memory. When fine-tuning the proposed method, we choose the easy, hard, and joint training sets of CMOTB, respectively, and optimized via AdamW with an initial learning rate of 8e-5. The model is trained for 40 epochs with a batch size of 8, each epoch contains 10,000 samples.  All other training parameters are consistent with the baseline~\cite{artrackv2}. We utilize the ViT-base as the backbone, and the total parameters are about 173M. The input resolution of the search region is $256 \times 256$. The model achieves real-time performance of over 65 FPS.

\begin{figure*}[htbp]
\centering
  \includegraphics[width=0.8\textwidth]{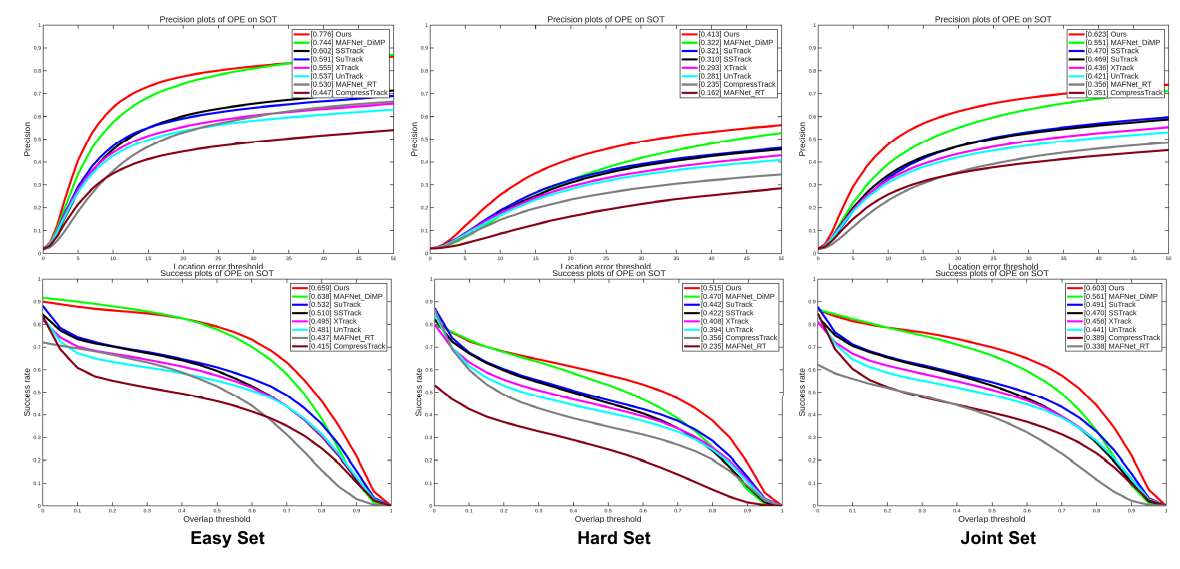}   
  \caption{The performance of different trackers on the CMOTB testing set.} 
  \label{fig:visual}  
\end{figure*}
\begin{table*}
    \centering
    \setlength{\tabcolsep}{2pt}
    \small
        \caption{The performance of various trackers on 18 challenging attributes using the CMOTB easy testing set is evaluated in terms of SR score. The best results are highlighted in \textbf{bold}.}
    \resizebox{\linewidth}{!}{\begin{tabular}{ccccccccccccccccccccc}
        \toprule
        &Trackers & FO & PO & DEF & SV & ROT & FM & CM & IV & TB & OV & BC & SIO & LR & ARC & VC & MA & MM & MD & \textbf{ALL} \\
        \midrule
        \multirow{8}{*}{\rotatebox{90}{Single}} 
        &MDNet~\cite{mdnet} & 38.9 & 41.0 & 31.5 & 41.9 & 42.1 & 35.8 & 42.7 & 42.2 & 38.5 & 41.2 & 39.5 & 40.9 & 41.1 & 42.1 & 35.8 & 41.4 & 46.3 & 37.7  & 42.7\\
        &SiamMask~\cite{siammask}& 39.7 & 43.8 & 39.1 & 45.1 & 45.7 & 37.8 & 44.9 & 45.9 & 39.4 & 51.9 & 44.2 & 43.7 & 42.1 & 45.0 & 38.4 & 46.6 & 47.5 & 44.4 & 45.7  \\
        &ATOM~\cite{atom} & 42.3 & 44.9 & 47.1 & 47.6 & 48.0 & 45.0 & 48.0 & 49.8 & 40.6 & 46.4 & 46.1 & 43.7 & 46.5 & 46.9 & 40.5 & 49.4 & 48.0 & 49.1 & 48.3 \\
        &TransT~\cite{transt} & 48.8 & 48.7 & 53.1 & 50.6 & 51.2 & 45.2 & 51.0 & 51.5 & 42.9 & 54.7 & 51.2& 47.0 & 54.4 & 48.9 & 44.4 & 51.2 & 52.7 & 50.7 & 51.0  \\
        &DiMP~\cite{dimp} & 46.9 & 48.3 & 51.6 & 50.8 & 50.7 & 48.0 & 50.7 & 52.6 & 43.5 & 55.6 & 49.8 & 46.7 & 55.2 & 50.9 & 44.6 & 53.6 & 52.9 & 51.0 & 51.3  \\
        &Stark~\cite{stark} & 48.2 & 47.6 & 50.9 & 50.2 & 48.7 & 43.6 & 50.6 & 52.8 & 43.5 & 52.0 & 50.4 & 44.7 & 49.4 & 48.6 & 46.5 & 52.1 & 52.1 & 53.4 & 50.4  \\
        &TrDimp~\cite{trdimp} & 52.4 & 52.3 & 50.9 & 53.6 & 51.8 & 46.1 & 53.2 & 54.2 & 47.5 & 57.4 & 52.3 & 52.5 & 53.7 & 53.8 & 46.6 & 54.1 & 56.8 & 51.2 & 53.9  \\
        &MixFormerV2~\cite{mixformerv2} & 44.6 & 46.0 & 49.2 & 47.6 & 47.2 & 44.4 & 47.8 & 48.8 & 41.1 & 53.3 & 45.8 & 44.9 & 49.8 & 46.7 & 42.8 & 48.3 & 49.6 & 49.2 & 47.8  \\       
        \midrule
        \multirow{3}{*}{\rotatebox{90}{Cross}}&$\mathrm{MArMOT}_{d}$~\cite{marmot} & 58.8 & 61.8 & 59.5 & 63.5 & 63.9 & 60.1 & 63.3 & 64.1 & 58.6 & 66.3 & 62.6 & 59.7 & 66.6 & 62.1 & 58.0 & 65.7 & 66.8 & 64.8 & 63.6  \\
        &$\mathrm{MAFNet}_d$~\cite{cmotb} & 58.7 & 61.6 & 54.7 & 63.7 & 64.1 & \textbf{62.3} & 63.8 & 64.3 & \textbf{60.4} & 66.1 & 62.3 & 58.9 & \textbf{66.7} & 61.9 & 57.4 & 65.7 & 66.6 & 64.8 & 63.8  \\
           &\cellcolor{gray!20}\textbf{Ours} & \cellcolor{gray!20}\textbf{58.9} & \cellcolor{gray!20}\textbf{64.6} & \cellcolor{gray!20}\textbf{66.0} & \cellcolor{gray!20}\textbf{65.7} & \cellcolor{gray!20}\textbf{66.3} & \cellcolor{gray!20}60.6 & \cellcolor{gray!20}\textbf{65.4} & \cellcolor{gray!20}\textbf{66.7} & \cellcolor{gray!20}59.5 & \cellcolor{gray!20}\textbf{68.6} & \cellcolor{gray!20}\textbf{64.5} & \cellcolor{gray!20}\textbf{60.9} &\cellcolor{gray!20} 52.9 & \cellcolor{gray!20}\textbf{64.1} & \cellcolor{gray!20}\textbf{62.4} & \cellcolor{gray!20}\textbf{67.6} &\cellcolor{gray!20}\textbf{69.3} & \cellcolor{gray!20}\textbf{67.6} & \cellcolor{gray!20}\textbf{65.9}  \\
         \bottomrule
    \end{tabular}}
    \label{tab:tracking_performance}
\end{table*}
\begin{figure}[htbp]
\centering
  \includegraphics[width=0.45\textwidth]{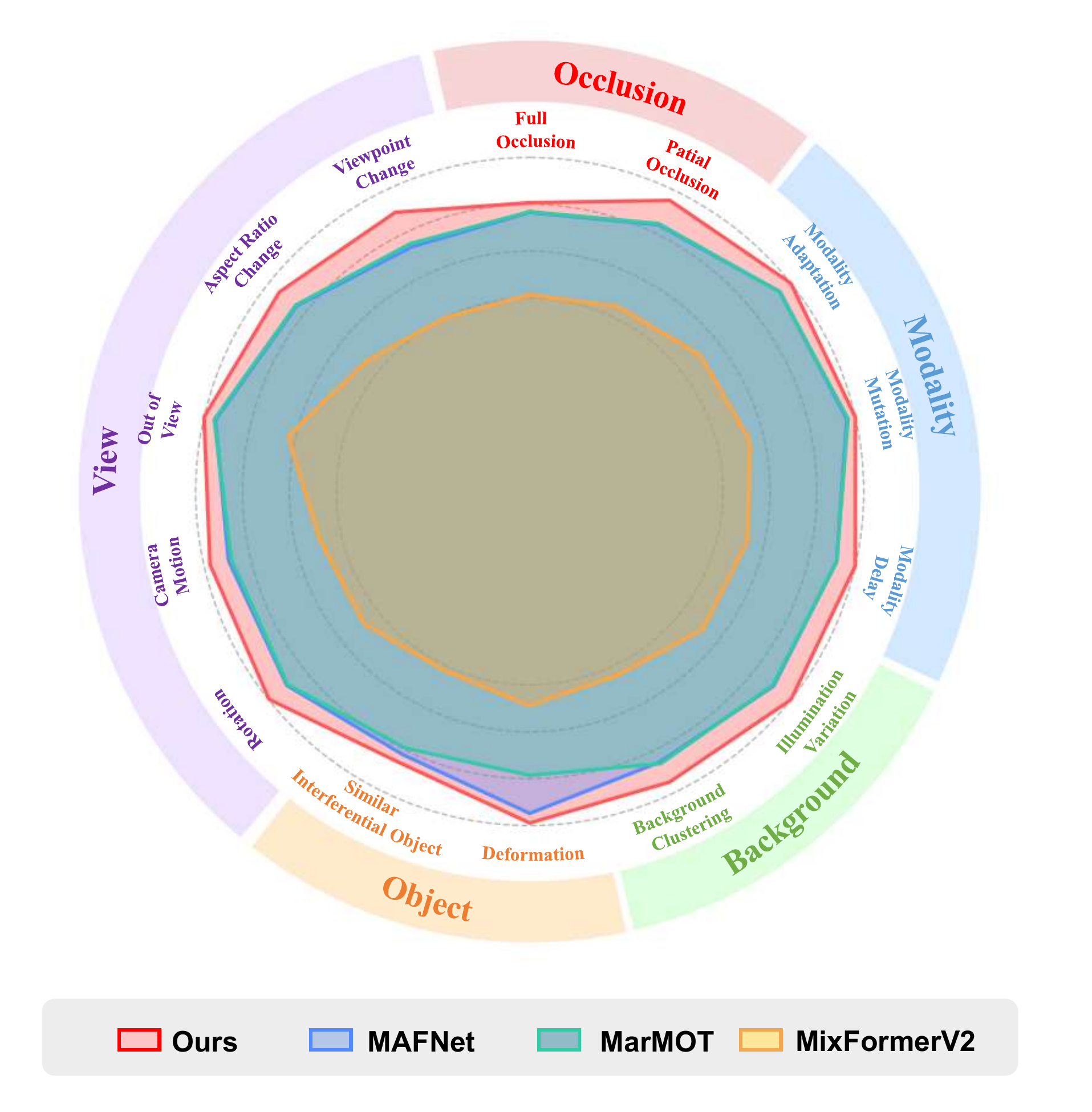}   
  \caption{Comparison of PR on different attributes in CMOTB.} 
  \label{fig:radar}  
\end{figure}
\begin{table}
    \centering
    \small
    \setlength{\tabcolsep}{3pt}
    \caption{Component analysis on CMOTB joint set, NGA represents NIR gated adapter, MCL represents modality consistent loss, and CTP represents consistent trajectory prediction. The best results are highlighted in \textbf{bold}.}
    \label{tableabla}
    \begin{tabular}{ccc|cccc|cc}
        \toprule
       NGA & MCL & CTP &   PR$\uparrow $ &$\Delta\%$ & SR$\uparrow$&$\Delta\%$ & FPS$\uparrow$ & Params\\
        \midrule
&  & & 0.581&- & 0.575&-&  \textbf{95}& 135M \\
\ding{51} & &  & 0.605&\textcolor{green}{+2.4} & 0.592&\textcolor{green}{+1.7}& 70& 173M \\
& \ding{51} & & 0.605 &\textcolor{green}{+2.4}& 0.587&\textcolor{green}{+1.2}&\textbf{95}& 135M\\
\ding{51}& \ding{51} & & 0.614&\textcolor{green}{+3.3} & 0.595&\textcolor{green}{+2.0}&70& 173M\\
 \rowcolor{gray!20}\ding{51}& \ding{51} &\ding{51} & \textbf{0.623}&\textcolor{green}{+4.2} & \textbf{0.604}&\textcolor{green}{+2.9}&  65& 173M\\
        \bottomrule
    \end{tabular} 
\end{table}
\subsection{Comparison with State-of-the-Arts}
\textbf{Quantitative Comparison.}
To validate the effectiveness of our method, we compare it with a range of state-of-the-art trackers:  TransT~\cite{transt}, OSTrack~\cite{ostrack}, AiATrack~\cite{aiatrack}, ARTrack~\cite{ARTRACK}, SeqTrack~\cite{seqtrack}, DropTrack~\cite{dropTRACK}, MixFormerV2~\cite{mixformerv2}, GRM~\cite{GRM}, RomTrack~\cite{ROMTRACK}, SSTrack~\cite{SSTrack}, CompresTrack~\cite{compresstrack}  SDSTrack~\cite{sdstrack}, ViPT~\cite{VIPT}, BAT~\cite{bat}, UNTrackr~\cite{untrack}, SuTrack~\cite{sutrack} and XTrack~\cite{xtrack},  MarMOT~\cite{marmot} and MAFNet~\cite{cmotb}. The results of other trackers are referred from benchmark~\cite{cmotb,prototype}
All methods were evaluated on the CMOTB test set, with comparisons conducted across three subsets: Easy Set, Hard Set, and Joint Set. As shown in Table~\ref{tablecompare}, the proposed method achieves the best performance across all metrics. Notably, compared with the best-performing single-modal method, DropTrack, our method achieves significant improvements of 5.4\% in PR and 4.7\% in SR on the hard subset. This considerable gain primarily stems from the limitation of single-modal trackers, which rely on a unified network to process RGB and NIR frames without specifically modeling modality-specific features, thus compromising tracking robustness.

Compared to the leading multi-modal method, BAT, our tracker achieves improvements of 15.2\% in PR and 14.6\% in SR on the hard subset. This is because multi-modal trackers heavily rely on paired dual-modal inputs and focus primarily on feature fusion. However, in cross-modal tracking, only a single modality is available at a time, rendering these fusion strategies less effective.
Furthermore, against the strongest cross-modal tracking baseline, our method still demonstrates substantial improvements of 7.2\% in PR and 4.3\% in SR overall. These gains are mainly attributed to our state switch that adaptively handles RGB, NIR, and invalid states, the NIR-Adapter, which refines NIR features specifically, and the proposed CTP module that ensures robust localization during invalid states.

\textbf{Attribute-based Analysis.}
To further demonstrate the effectiveness of our proposed method, we conduct attribute-based comparison experiments against the SOTA methods across 18 key challenge attributes. 
These attributes include full occlusion (FO), partial occlusion (PO), deformation (DEF), scale variation (SV), rotation (ROT), fast motion (FM), camera motion (CM), illumination variation (IV), target blur (TB), out-of-view (OV), background clustering (BC), similar interferential object (SIO), low resolution (LR), aspect ratio change (ARC), viewpoint change (VC), modality adaptation (MA), modality mutation (MM), and modality delay (MD). 
Following previous works, we use the success rate metric for evaluation, with results presented in Table~\ref{tab:tracking_performance} and Figure~\ref{fig:radar}.
Our method outperforms the best existing approach by 2.1\% in overall SR and achieves consistent improvements across most challenge attributes. Specifically, for modality-related challenges, MA, MM, and MD, our approach improves SR by 1.2\%, 3\%, and 1.3\%. Additionally, for general tracking challenges such as FO, PO, and ARC, our method surpasses the best baseline by 1.9\%, 2.7\%, and 2.8\%, demonstrating its robustness against common difficulties like occlusion. The only slight performance drop occurs in TB, FM, and LR attributes compared to the best method.

Overall, these results provide evidence supporting the rationality
and the effectiveness of our proposed method, demonstrating its ability
to enhance robustness
against challenging conditions.

\textbf{Cross-dataset Generalization.} To further assess generalization, we conduct a cross-dataset experiment despite the limited availability of cross-modal benchmarks. Beyond CMOTB, we identify 21 cross-modal sequences in TNL2K that are RGB–T rather than our training setting RGB-NIR. Given the small sample size, we perform zero-shot evaluation directly using the CMOTB-trained weights without any fine-tuning and obtain +1.2\%/+0.8\% improvement over the baseline under the same resolution/runtime settings, indicating that our method retains non-trivial cross-modal generalization even without adaptation to the target modality.

\begin{table}
    \centering
    \setlength{\tabcolsep}{7pt}
    \small
    \caption{Comparison between different motion estimation methods on the CMOTB joint set. The best results are highlighted in \textbf{bold}.}
    \label{tableablamo}
    \resizebox{1\linewidth}{!}{\begin{tabular}{c|cccc}
        \toprule
       Settings &   PR$\uparrow$&$\Delta\%$ & SR$\uparrow$&$\Delta\%$\\
        \midrule
w/o Motion Estimation&  0.614 &-& 0.595&- \\
Kalman Filter&  0.618 &\textcolor{green}{+0.4} &0.599&\textcolor{green}{+0.4} \\
Extended Kalman Filter & 0.620 &\textcolor{green}{+0.6}& 0.600 &\textcolor{green}{+0.5}\\
 \rowcolor{gray!20}Consistence Trajectory Prediction & \textbf{0.623} &\textcolor{green}{+0.9}& \textbf{0.604} &\textcolor{green}{+0.9}\\
        \bottomrule
    \end{tabular} }
\end{table}
\begin{table}[htbp]
\centering
\small
\setlength{\tabcolsep}{13pt}
\caption{Threshold sensitivity for the invalid-ratio on CMOTB joint set. The best results are
highlighted in bold.}
\label{tab:threshold_sensitivity}
\begin{tabular}{@{}rccccc@{}}
\toprule
\textbf{$\rho$ (\%)} & 20 & 30 & \textbf{40 (Ours)} & 50 & 60 \\
\midrule
\textbf{PR$\uparrow$} & 60.8 & 61.9 & \textbf{62.3} & 62.3 & 61.7 \\
\textbf{SR$\uparrow$} & 59.2 & 60.0 & \textbf{60.4} & 60.2 & 59.9 \\
\bottomrule
\end{tabular}
\end{table}
\subsection{Ablation Study}
\textbf{Component Analysis.}
To further validate the effectiveness of each component in our method, we conducted an ablation study on the joint set of the CMOTB dataset, with results presented in Table~\ref{tableabla}. The Adapter represents the NIR gated adapter, Loss represents the similarity loss, and CTP represents consistent trajectory prediction. 

The first row of Table~\ref{tableabla} shows the performance of the baseline method, which achieves 58.1\% PR and 57.5\% SR, using a unified encoder without any modality-specific design. In the second row, we introduce our proposed NIR gated adapter into the baseline. This module specifically refines NIR features by referencing dynamic templates, leading to substantial improvements of 2.4\% in PR and 1.7\% in SR. Although it introduces a slight drop in inference speed, the notable gains in performance clearly demonstrate the effectiveness of adapting to modality-specific characteristics. In the fourth row, we replace the baseline loss function with our feature similarity loss, which promotes cross-modal consistency by aligning RGB and NIR feature distributions. This adjustment brings an additional 0.9\% improvement in PR and 0.3\% in SR, demonstrating the value of explicitly optimizing feature similarity across modalities.
Finally, in the last row, we incorporate our CTP module, designed to predict under an invalid state. This further boosts both PR and SR by 0.9\%, showing its effectiveness in mitigating performance degradation under invalid state.

Overall, these results provide evidence supporting the rationality and effectiveness of our proposed method, demonstrating its ability to enhance modality adaptation, feature alignment, and robustness against challenging conditions.

\begin{figure*}[t]
\centering
  \includegraphics[width=0.9\textwidth]{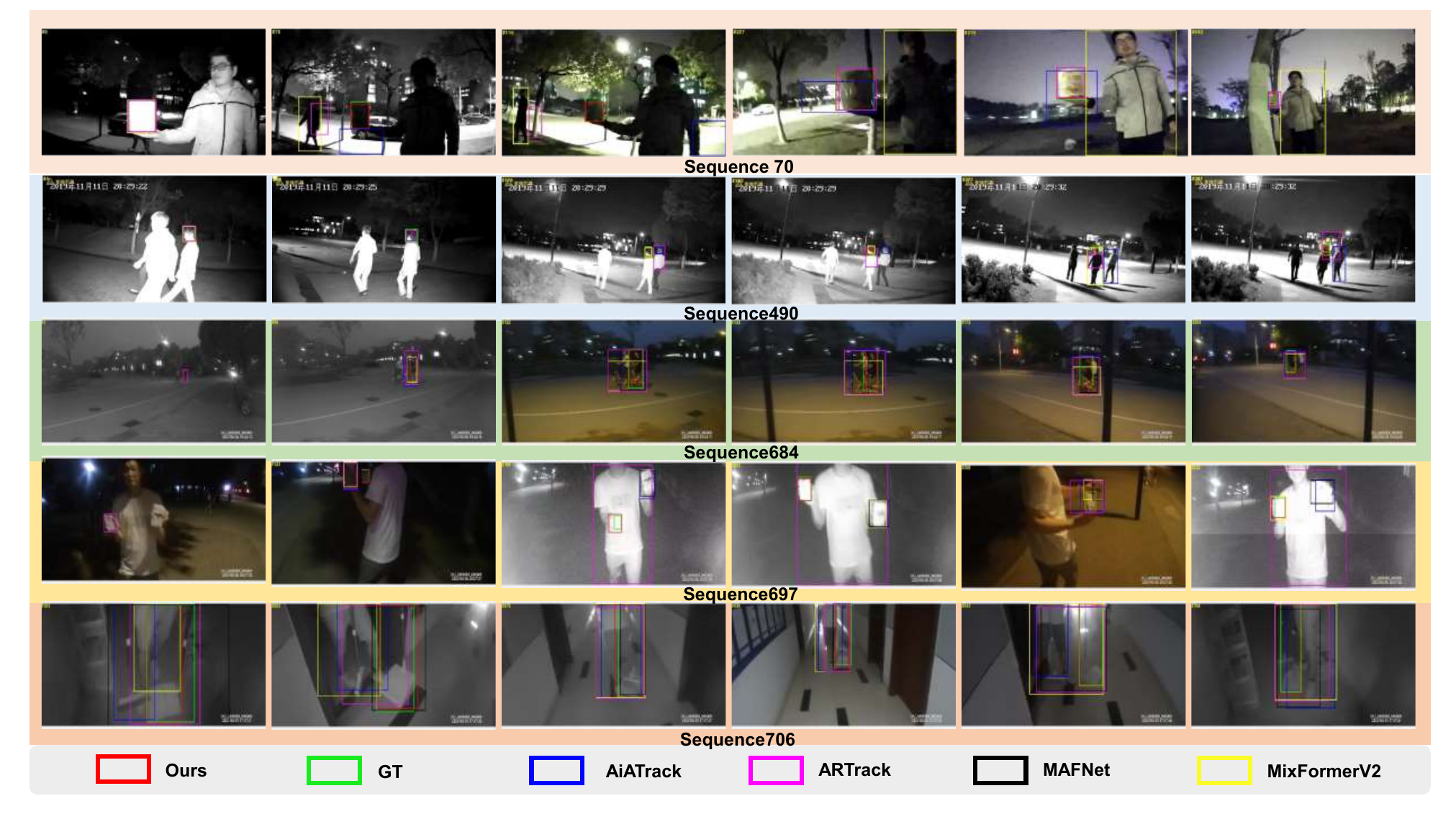}   
  \caption{Qualitative comparison between the proposed method and state-of-the-art methods on the CMOTB test set, the ground truth is marked in green.} 
  \label{fig:visual}  
\end{figure*}

\begin{figure*}[t]
\centering
  \includegraphics[width=0.9\textwidth]{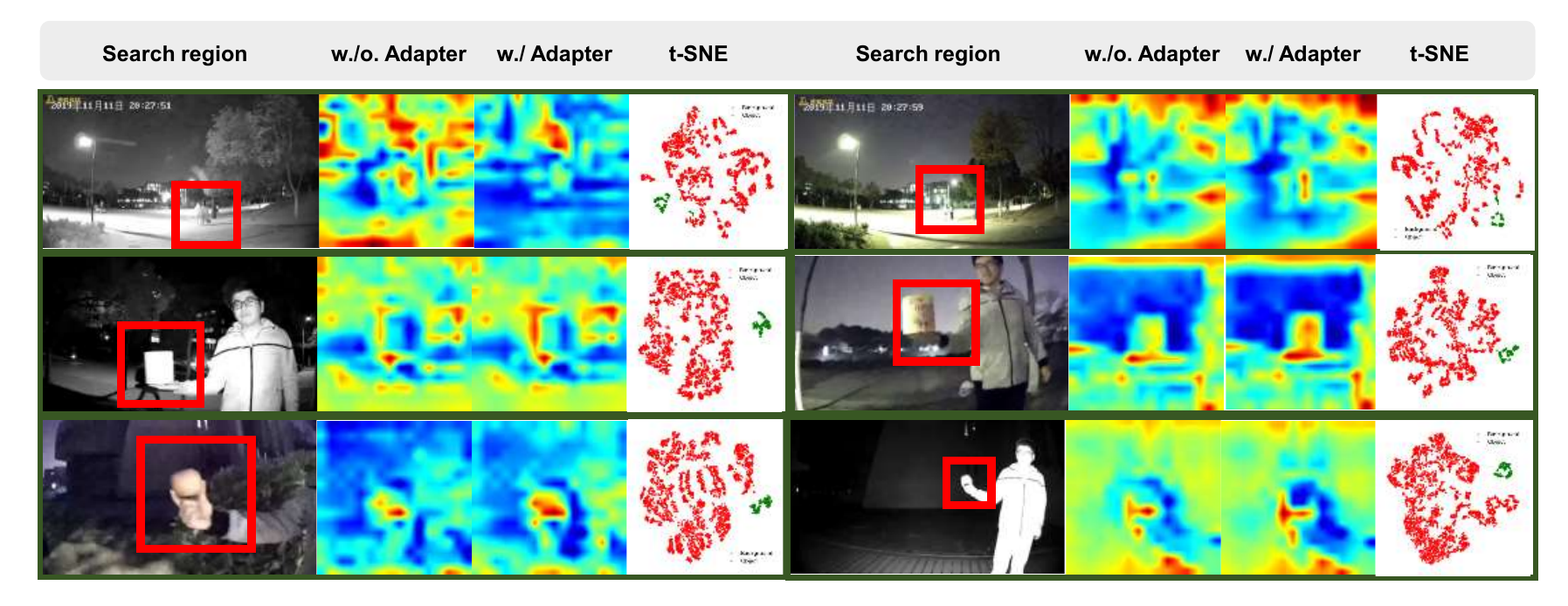}   
  \caption{Visualization results highlighting the impact of the NIR gated adapter on feature representations.} 
  \label{fig:visuafeat}  
\end{figure*}
\begin{figure}[t]
\centering
  \includegraphics[width=0.5\textwidth]{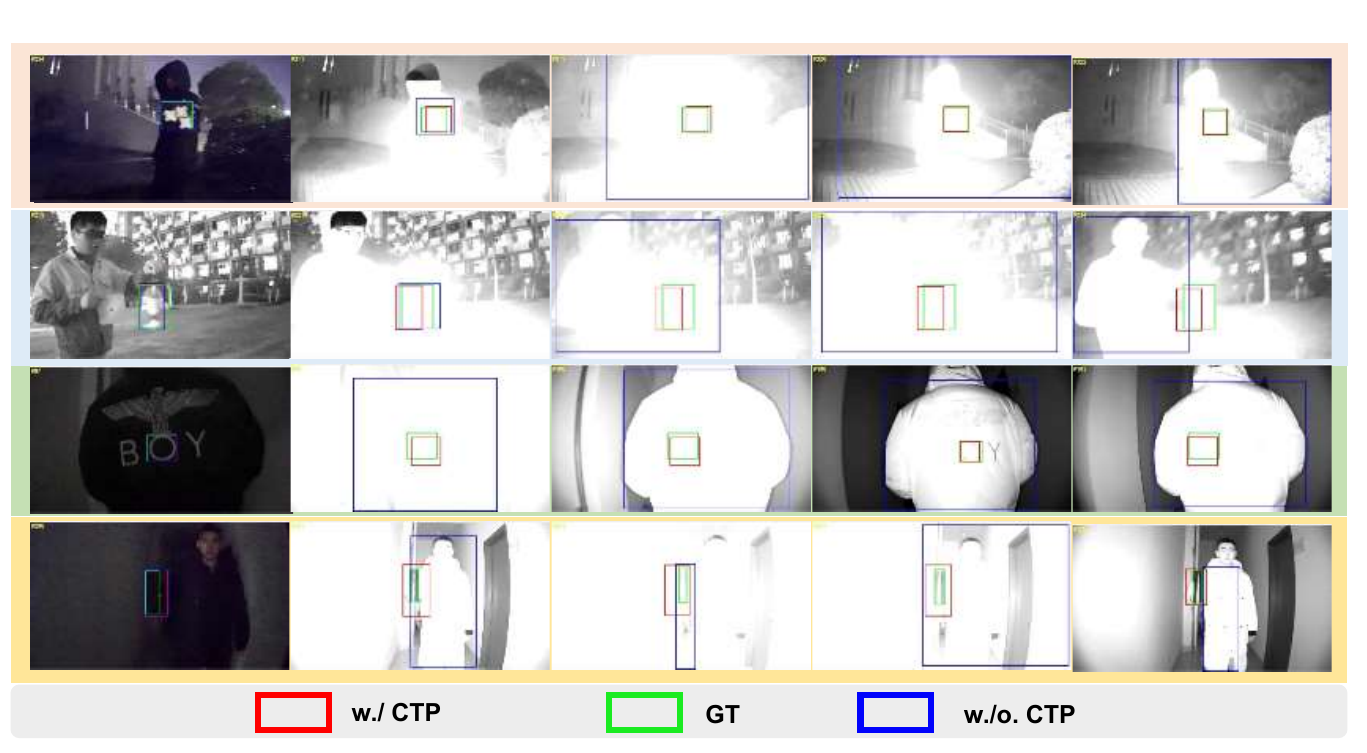}   
  \caption{Visualization results of invalid state in cross-modal tracking with CTP module.} 
  \label{fig:visualab}  
\end{figure}
\begin{figure}[t]
\centering
  \includegraphics[width=0.5\textwidth]{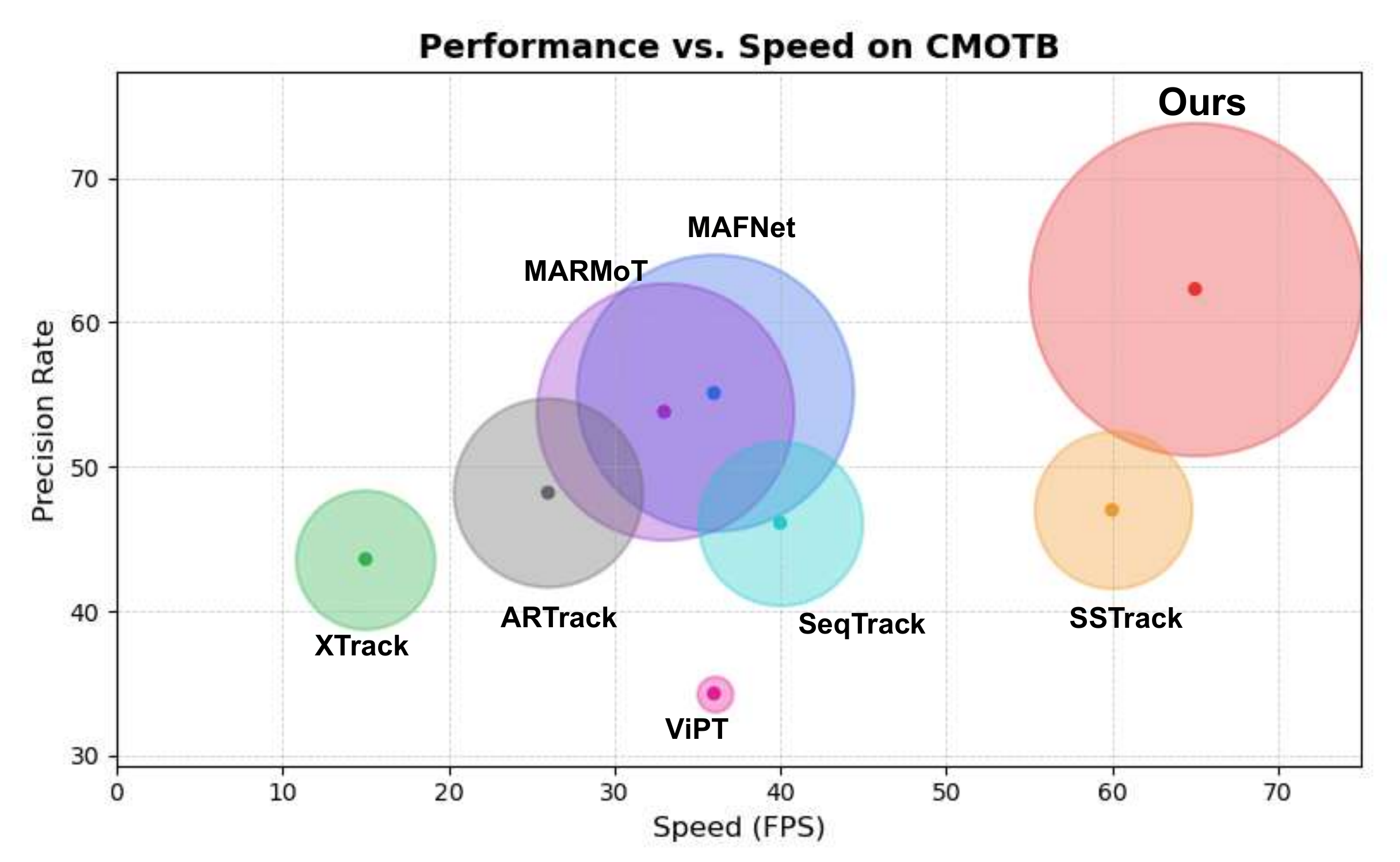}   
  \caption{Comparison of PR vs. FPS for different tracking methods in CMOTB.} 
  \label{fig:bubble}  
\end{figure}
\begin{figure}[t]
\centering
  \includegraphics[width=0.5\textwidth]{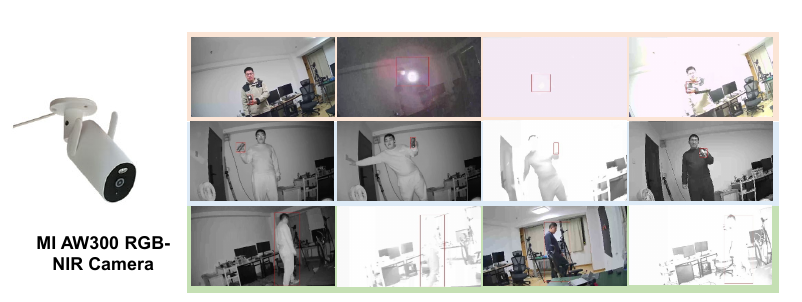}   
  \caption{Visualization results of real-world test.} 
  \label{fig:real}  
\end{figure}
\textbf{Comparison of Motion Estimation.}
To validate the effectiveness of our proposed CTP module, we conduct comparative experiments against several commonly adopted motion estimation methods in visual tracking. As illustrated in Table~\ref{tableablamo}, the first row shows the performance of the baseline model without any motion estimation module, achieving 61.4\% in PR and 59.5\% in SR.
In the second row, we introduce a Kalman filter to handle invalid modality states, which brings a 0.4\% improvement in both PR and SR. This clearly demonstrates the necessity of incorporating motion estimation when the modality becomes unreliable.
To address the non-linearity often encountered in real-world tracking, we further replace the Kalman filter with a non-linear extended Kalman filter, which yields an additional 0.2\% and 0.1\% gain in PR and SR, respectively. Finally, we integrate our CTP module, which leverages both modality and tracking reliability to adaptively update the prediction model. This leads to a further improvement of 0.3\% in PR and 0.4\% in SR, achieving a total gain of 0.9\% in both metrics over the baseline.

These results not only highlight the necessity of motion estimation under the invalid state but also demonstrate the effectiveness of our modality-aware design in cross-modal tracking scenarios.

\textbf{Parameter Analysis.}
We evaluate the impact of key parameters and the robustness of our method via an ablation on the invalid-ratio threshold used to detect modality failure. Specifically, we vary the threshold from 20\% to 60\% and test on the joint set. As shown in Table~\ref{tab:threshold_sensitivity}, performance fluctuates only mildly across this range, indicating that our method is not sensitive to the threshold and is robust overall. The best results occur at 40\%, likely because overly small or overly large thresholds misjudge failure states and, in turn, adversely affect subsequent tracking.

\textbf{Discussion.} Tracking failures often occur during invalid modality states, where the visual features become unreliable due to overexposure or sensor limitations. This issue leads to target drift, especially when trackers rely solely on visual input for target localization. In response, we propose a novel approach using the Extended Kalman Filter~\cite{ekf} to predict the target’s trajectory during such invalid states.

\paragraph{Kalman Filters and their Limitations in Cross-Modal Tracking}
While traditional Kalman filters have been applied in tracking tasks, including handling occlusions and associating trajectories~\cite{kalman}, they are typically limited by their reliance on linear motion models. These models are ill-suited for tracking objects in dynamic environments, where motion patterns are inherently nonlinear. In such settings, the Kalman filter struggles to accurately predict trajectories, resulting in poor performance during non-linear motion transitions, such as fast changes in direction or speed.

\paragraph{Extended Kalman Filter for Nonlinear Motion}
To address the inherent nonlinearities in cross-modal tracking, we extend the standard Kalman filter to the Extended Kalman Filter. Unlike its linear counterpart, EKF can better approximate nonlinear systems by iteratively linearizing the state model at each time step. This makes EKF more robust in handling complex motion dynamics, which are common in cross-modal tracking when the modality switches frequently.

\paragraph{Reliability Estimation in Invalid States}
A further challenge arises when unreliable visual features persist across multiple frames during invalid modality states. Standard EKF assumes consistent updates from visual observations, which is problematic when the input features become unstable. In such cases, naive corrections may accumulate, causing significant errors in the predicted trajectory. 

To overcome this, we introduce an adaptive mechanism within EKF. We incorporate reliability estimation based on the modality's state (\emph{e.g.}, whether it is RGB, NIR, or invalid) and tracking confidence scores. This adaptive approach dynamically adjusts the EKF’s noise covariance matrices, reducing the influence of unreliable observations. As a result, the tracker becomes more resilient to erroneous inputs and can maintain accurate trajectory predictions during invalid states, leading to more stable and robust cross-modal tracking performance.


\subsection{Visualization Analysis}
\textbf{Tracking Result Visualization.} 
We provide qualitative comparisons between SwiTrack and state-of-the-art trackers under challenging scenes that are typical for cross-modal switches, illumination over-exposure, distractor objects with similar appearance, heavy occlusions, and fast motion, as shown in Figure.~\ref{fig:visual}, as commonly observed on CMOT-style data, modality may switch multiple times within the same sequence, and the visible stream can momentarily collapse, stressing temporal consistency and modality awareness. Across representative sequences, SwiTrack maintains stable localization in both RGB and NIR conditions, preserves tighter boxes near occlusion boundaries, and is less prone to drift when distractors or exposure spikes appear, reflecting the benefit of explicit state switching and reliability-weighted prediction. 


\textbf{Consistent Trajectory Prediction Visualization.}
To intuitively demonstrate the effectiveness of our proposed CTP module in handling invalid states, we provide qualitative visualizations in Figure~\ref{fig:visualab}. As shown, when the CTP module is not employed, the tracker struggles to localize the target once the input modality becomes invalid, typically due to abrupt changes. The failure leads to a rapid loss of the target and degrades tracking performance in subsequent frames. In contrast, when the consistent trajectory prediction module is enabled, the tracker leverages motion cues to predict the target’s position during an invalid state, successfully maintaining accurate localization.

\textbf{Feature Visualization.}
To provide a more intuitive demonstration of the effectiveness of our method, we conduct feature visualization, as shown in Figure~\ref{fig:visuafeat}. The first column displays the original frames along with the corresponding search regions. The second column presents the feature maps generated without the use of the NIR-Adapter, where both RGB and NIR features are uniformly processed by the shared visual encoder. The third column illustrates the feature maps from our full method, where the NIR-Adapter is applied on top of the visual encoder to specifically refine the NIR features. The fourth column illustrates the t-SNE map of features, green points denote target features and red points denote background features, revealing a clear separation achieved by our method.
As can be observed, our method produces more discriminative and concentrated target representations in both RGB and NIR scenarios. This demonstrates that our design not only preserves the shared representations but also strengthens modality-specific features, thereby improving the overall feature quality for robust cross-modal tracking.
\subsection{Real World Test}
To further assess practical capability of SwiTrack, we conduct a real-world test using a cross-modal camera MI AW300 that automatically switches between visible and near-infrared modes and activates an auxiliary illuminator under low light. As shown in Figure~\ref{fig:real}, the sequences cover diverse scenes and include visible, NIR, and invalid states. Across all conditions, our method consistently maintained stable tracking, demonstrating strong robustness and real-world applicability.
\subsection{Performance Efficiency Analysis}
To compare our method with SOTA trackers more intuitively, we conduct a performance efficiency study, as shown in Figure~\ref{fig:bubble}, the x-axis denotes model speed, the y-axis denotes PR, and the bubble area represents SR. Bubbles closer to the upper-right and with larger areas indicate better overall performance. As shown, SwiTrack sits in the upper-right with the largest bubble, outperforming others in both tracking speed and accuracy.
\section{Conclusion}

In this paper, we proposed SwiTrack, a novel cross-modal tracker that categorizes input into three states: RGB, NIR, and invalid. To effectively handle state variations, we proposed the NIR gated adapter in conjunction with a unified visual encoder, which adaptively refines NIR features, while RGB frames are processed directly by the encoder. Additionally, we designed a consistency trajectory prediction module to predict target trajectories during an invalid state, thereby boosting tracking accuracy. Extensive experiments on recent cross-modal tracking benchmarks demonstrate that our method achieves state-of-the-art performance.

\bibliographystyle{IEEEbib}
\bibliography{mybib}


 


\begin{IEEEbiography}[{\includegraphics[width=1in,height=1.25in,clip,keepaspectratio]{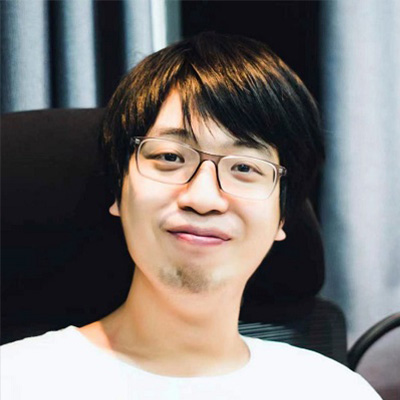}}]{Boyue Xu} (Student Member, IEEE) received the M.S. degree from the School of Software Engineering at Nanjing University, Nanjing, China, where he is currently pursuing the Ph.D. degree at the School of Department of Computer Science and Technology, Nanjing University. His current research interests include multi-modal detection and tracking. 
\end{IEEEbiography}

\begin{IEEEbiography}[{\includegraphics[width=1in,height=1.25in,clip,keepaspectratio]{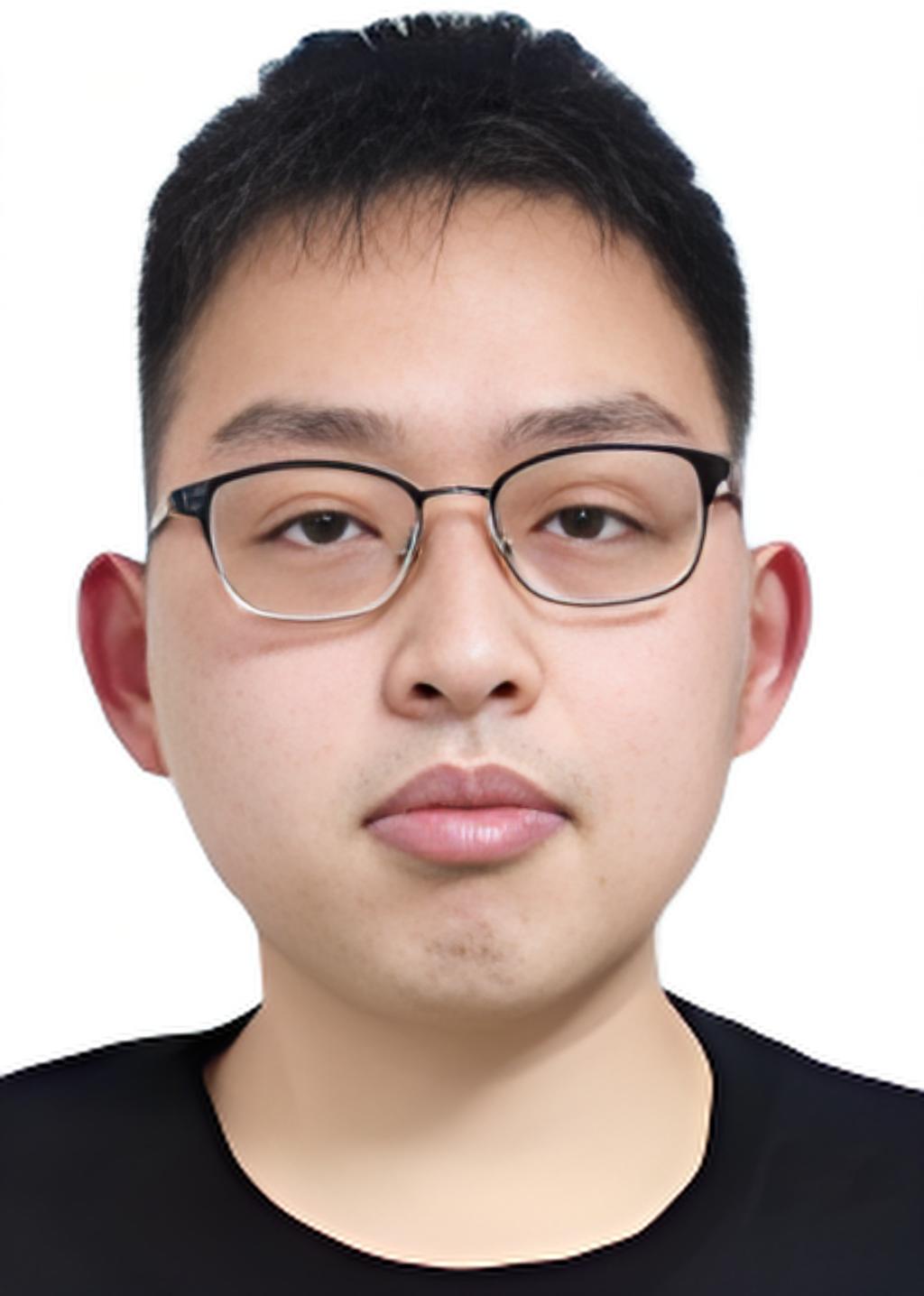}}]{Ruichao Hou} (Member, IEEE) received his Ph.D degree from the Department of Computer Science and Technology, Nanjing University in 2023.
He is currently an Assistant Researcher at the Software Institute of Nanjing University. His research mainly focuses on multi-modal object detection and tracking. He has published more than 20 papers in top-tier journals and conferences. 
\end{IEEEbiography}

 \begin{IEEEbiography}[{\includegraphics[width=0.9in,height=1.25in,clip,keepaspectratio]{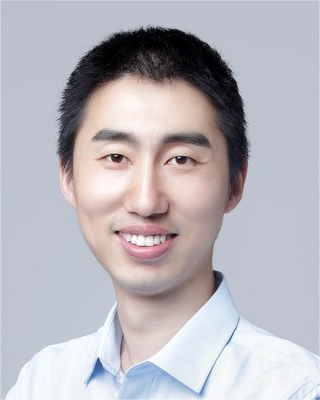}}] {Tongwei Ren} (Member, IEEE) received the B.S., M.E., and Ph.D. degrees from Nanjing University, Nanjing, China, in 2004, 2006, and 2010, respectively. He joined Nanjing University in 2010, and at present he is a professor. His research interest mainly includes multimedia computing and its real-world applications. He has published more than 40 papers in top-tier journals and conferences. He was a recipient of the best paper candidate awards of ICIMCS 2014, PCM 2015, and MMAsia 2020, and he was in the champion teams of ECCV 2018 PIC challenge, MM 2019 VRU challenge, MM 2020 DVU challenge, MM 2022 DVU challenge and MM 2023 DVU challenge.
 \end{IEEEbiography}
 
\begin{IEEEbiography}[{\includegraphics[width=1in,height=1.25in,clip,keepaspectratio]{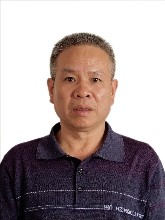}}] {Dongming Zhou} received the B.S. and M.S. degrees in industry automatization from the Department of Automatic Control Engineering, Huazhong University of Science and Technology, Wuhan, China, in 1985 and 1988, respectively, and the Ph. D. degree in circuitry
and system from the School of Information Science and Technology, Fudan University, Shanghai, China, in 2004. In 2008, he was a Visiting Scholar with York University, Toronto, ON, Canada. He is currently a Professor at the School of Information Science and Engineering, Yunnan University, Kunming, China. His current research interests include biomedical engineering, computational intelligence, complex systems, neural networks, and their applications.
 \end{IEEEbiography}

 \begin{IEEEbiography}[{\includegraphics[width=1in,height=1.25in,clip,keepaspectratio]{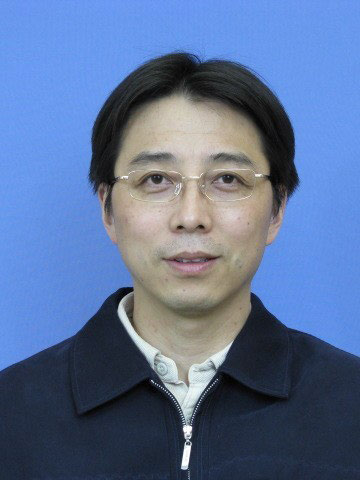}}] {Gangshan Wu} (Member, IEEE) received the B.Sc., M.S., and Ph.D. degrees from the Department of Computer Science and Technology, Nanjing University, Nanjing, China, in 1988, 1991, and 2000, respectively. He is currently a Professor with the School of Computer Science, Nanjing University. His current research interests include computer vision, multimedia content analysis, multimedia information retrieval, digital museum, and large-scale volumetric data processing.
 \end{IEEEbiography}

\begin{IEEEbiography}[{\includegraphics[width=1in,height=1.25in,clip,keepaspectratio]{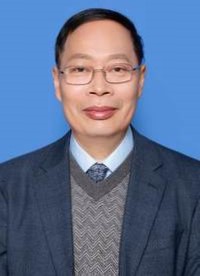}}]{Jinde Cao}(Fellow, IEEE) received the B.S. degree in mathematics from Anhui Normal University, Wuhu, China in 1986, the M.S. degree from Yunnan University, Kunming, China, and the Ph.D. degree from Sichuan University, Chengdu, China, both in applied mathematics, 1989, and 1998, respectively. He was a Postdoctoral Research Fellow at the Department of Automation and Computer-Aided Engineering, Chinese University of Hong Kong, Hong Kong, China from 2001 to 2002.

Professor Cao is an Endowed Chair Professor, the Dean of Science Department and the Director of the Research Center for Complex Systems and Network Sciences at Southeast University (SEU). He is also the Director of the National Center for Applied Mathematics at SEU-Jiangsu of China and the Director of the Jiangsu Provincial Key Laboratory of Networked Collective Intelligence of China. He is also Honorable Professor of Institute of Mathematics and Mathematical Modeling, Almaty, Kazakhstan.

Prof. Cao was a recipient of the National Innovation Award of China, IETI Annual Scientific Award, Obada Prize and the Highly Cited Researcher Award in Engineering, Computer Science, and Mathematics by Clarivate Analytics. He is elected as a member of Russian Academy of Sciences, a member of the Academia Europaea (Academy of Europe), a member of Russian Academy of Engineering, a member of the European Academy of Sciences and Arts, a member of the Lithuanian Academy of Sciences, a fellow of African Academy of Sciences, and a fellow of Pakistan Academy of Sciences.
\end{IEEEbiography}


\end{document}